\pdfoutput=1
\documentclass[10pt,twocolumn,letterpaper]{article}

\usepackage{wacv}

\newif\ifaddheader
\addheaderfalse
\def\addconfheader{\global\addheadertrue}

\addconfheader % *** Uncomment this line to add conference headers on every page

\usepackage{times}
\usepackage{epsfig}
\usepackage{graphicx}
\usepackage{amsmath}
\usepackage{amssymb}
\usepackage{multirow}
\usepackage[inline]{enumitem}
\usepackage{pifont}% http://ctan.org/pkg/pifont
\newcommand{\cmark}{\ding{51}}%
\newcommand{\xmark}{\ding{55}}%
\usepackage{cellspace}
\usepackage[noadjust]{cite}
\usepackage{mathtools}
\usepackage{esdiff}
\usepackage{cleveref}

\ifaddheader
\usepackage{fancyhdr} 
\fancypagestyle{titlestyle}
{
	\fancyhf{}
	\fancyfoot[C]{This paper is a preprint (IEEE “accepted” status at IEEE WACV 2019, Hawaii USA). IEEE copyright notice. \copyright~2018
			IEEE. Personal use of this material is permitted. Permission from IEEE must be
			obtained for all other uses, in any current or future media, including
			reprinting/republishing this material for advertising or promotional purposes,
			creating new collective works, for resale or redistribution to servers or lists, or
			reuse of any copyrighted}
}
\fi

\wacvfinalcopy % *** Uncomment this line for the final submission

\def\wacvPaperID{343} % *** Enter the wacv Paper ID here

\ifwacvfinal\fi % no headers or footers

\ifaddheader

\fi

\begin{document}

%%%%%%%%% TITLE
\title{Scalable Logo Recognition using Proxies}

%Authors at the same institution, not sure how to do the email, this does not look right
%\author{Istv\'an Feh\'erv\'ari \hspace{2cm} Srikar Appalaraju \\
%{\tt\small istvanfe@amazon.com \hspace{2cm} srikara@amazon.com} \\
%Amazon Inc.
%}

% Authors at different institutions, this repeats shared affiliat
%\author{Istv\'an Feh\'erv\'ari\\
%{\tt\small istvanfe@amazon.com}
%\and
%Srikar Appalaraju \\
%{\tt\small srikara@amazon.com}
%}

\author{
	\begin{tabular}[t]{c@{\extracolsep{8em}}c} 
		Istv\'an Feh\'erv\'ari  & Srikar Appalaraju \\
		{\tt\small istvanfe@amazon.com} & {\tt\small srikara@amazon.com} \\
		\multicolumn{2}{c}{Amazon Inc.}
	\end{tabular}
}

\maketitle
\ifwacvfinal\thispagestyle{empty}\fi
\ifaddheader\thispagestyle{titlestyle}\fi

%%%%%%%%% ABSTRACT
\begin{abstract}
	Logo recognition is the task of identifying and classifying logos. Logo recognition is a challenging problem as there is no clear definition of a logo and there are huge variations of logos, brands and re-training to cover every variation is impractical. In this paper, we formulate logo recognition as a few-shot object detection problem. The two main components in our pipeline are universal logo detector and few-shot logo recognizer. The universal logo detector is a class-agnostic deep object detector network which tries to learn the characteristics of what makes a logo. It predicts bounding boxes on likely logo regions. These logo regions are then classified by logo recognizer using nearest neighbor search, trained by triplet loss using proxies. We also annotated a first of its kind product logo dataset containing 2000 logos from 295K images collected from Amazon called \textit{PL2K}. Our pipeline achieves 97\% recall with 0.6 mAP on \textit{PL2K} test dataset and state-of-the-art 0.565 mAP on the publicly available FlickrLogos-32 test set without fine-tuning.
\end{abstract}

%-------------------------------------------------------------------------
%%%%%%%%% BODY TEXT
\section{Introduction}
\begin{figure}[t]
	\begin{center}		
		\setlength\cellspacetoplimit{4pt}
		\setlength\cellspacebottomlimit{4pt}
		
		\begin{tabular}{ Sc Sc Sc}
			\includegraphics[height=0.2\linewidth]{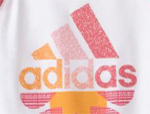} & \includegraphics[height=0.2\linewidth]{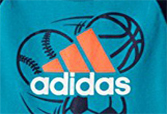} & \includegraphics[height=0.2\linewidth]{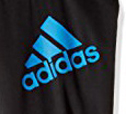} \\
			
			\includegraphics[height=0.2\linewidth]{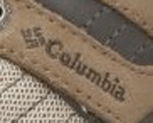} & \includegraphics[height=0.2\linewidth]{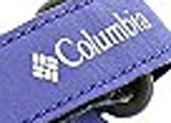} & \includegraphics[height=0.2\linewidth]{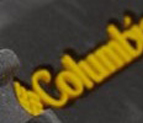} \\
			
			\includegraphics[height=0.15\linewidth]{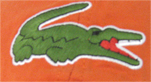} & \includegraphics[height=0.15\linewidth]{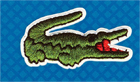} & \includegraphics[height=0.15\linewidth]{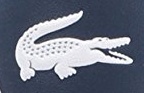} \\
			\hline
		\end{tabular}
		
		\begin{tabular}{ Sc Sc}
			
			\includegraphics[height=0.2\linewidth]{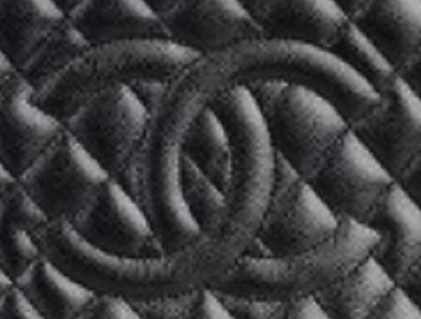} & \includegraphics[height=0.2\linewidth]{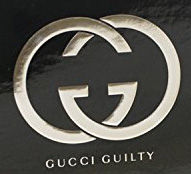} \\
			
			\includegraphics[height=0.2\linewidth]{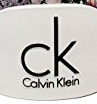} & \includegraphics[height=0.2\linewidth]{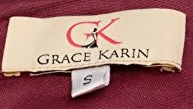} \\
			
			\includegraphics[height=0.2\linewidth]{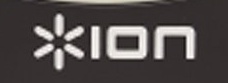} & \includegraphics[height=0.2\linewidth]{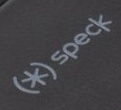} \\
			
		\end{tabular}
	\end{center}
	\caption{Logo variations exemplar images. (Row 1-3) Intra-class variations of brands Adidas, Columbia, Lacoste per row. Notice, different backgrounds, placement, fonts. (Row 4-6) Inter-class variations of brands Chanel - Gucci, Calvin Klein - Grace Karin, Ion - Speck. Notice, similar looking logos but belong to different brands.}
	\label{fig:example_logos}
	\label{fig:onecol}
\end{figure}

% covered problem definition, challenges and applications.
Logo recognition has a long history in Computer vision with works dating back to 1993 \cite{doermann1993logo}. While the problem is well defined (detect and identify brand logos in images), it is a challenging object recognition and classification problem as there is no clear definition of what constitutes a logo. A logo can be thought of as an artistic expression of a brand, it can be either a (stylized) letter or text, a graphical figure or any combination of these. Furthermore, some logos have a fixed set of colors with known fonts while others vary a lot in color and specialized unknown fonts. Additionally, due to the nature of a logo (as brand identity) there is no guarantee about its context or placement in an image, in reality logos could appear on any product, background or advertising surface. Also, this problem has large intra-class variations e.g. for a specific brand, there exist various logos types (old and new Adidas logos, small and big versions of Nike) and inter-class variations e.g. there exists logos which belong to different brands but look similar (see Figure \ref{fig:example_logos}).

Accurate logo recognition in images can have multiple applications. It can enable better semantic search, better personalized product recommendations, improved contextual ads, IP infringement detection amongst other applications. 

Logo recognition has many inter- and intra-class variations, retraining with each new variation is unscalable. In this work, we explore logo recognition as a few-shot problem. We show experimentally and empirically that our approach is able to detect and identify previously unseen (in training) logos. We show that our models have better learned what makes something a logo than prior art, with performance being the evidence. We created a first of its kind Product logo dataset \textit{PL2K}. It contains over 2000 logos with large inter- and intra-class variations. Our pipeline achieves 97\% recall with 0.6 mAP on the new \textit{PL2K} test dataset and state-of-the-art 0.565 mAP on the publicly available FlickrLogos-32 test set without fine-tuning. With this we present the main contributions of our work: 

\begin{itemize}
	\item Universal logo detector: a class-agnostic logo detector, capable of predicting bounding boxes on previously unseen logos.
	\item Novel logo recognizer: A network based on spatial transformer and proxy loss provides state-of-the-art results on FlickrLogos-32 test set. We further show by experiments that this type of architecture with metric learning does really well in few-shot logo recognition thereby providing a more generalized logo recognition model. 
	\item Product logo dataset \textit{PL2K} \footnote{An image with a single product in front of a white background}: We discuss how we went about collecting and annotating this large-scale logo dataset from the Amazon catalog.
\end{itemize}

\begin{figure}[t]
	\begin{center}
		\includegraphics[width=\linewidth]{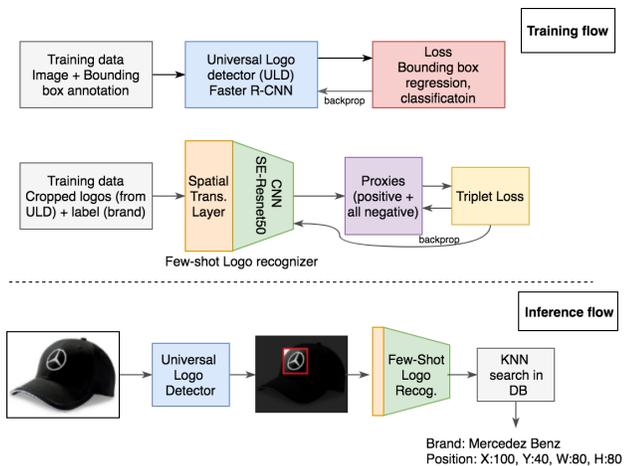}
	\end{center}
	\caption{Flow diagram for our Architecture for training and inference. We train proxies jointly with our few-shot model which are used to compute the triplet loss.}
	\label{fig:flow_architecture}
\end{figure}

%-------------------------------------------------------------------------
\section{Related works}
In this section we discuss closely related works in the fields of deep learning in computer vision, prior art in logo recognition, and metric learning.

\subsection {Deep learning}
Our work adopts on recent deep learning research state-of-the-art results in image object detection and classification \cite{rumelhart1986learning, krizhevsky2012imagenet, srivastava2014dropout, he2016deep, liu2016ssd, ioffe2015batch, girshick2015fast, redmon2016you, redmon2018yolov3}. The problem of logo recognition itself has a rich research history. In 1990's the problem was mainly explored in information retrieval use-cases. An image descriptor was generated using affine transformations and stored in a database for retrieval \cite{doermann1993logo}. There were also some neural network based approaches \cite{francesconi1997logo, cesarini1997neural} but the networks were not as deep nor the results as impressive as recent work.

In 2000's, with the advent of SIFT and related approaches \cite{lowe2004distinctive, bay2006surf, rublee2011orb} better image descriptors were possible. These methodologies were used to represent images better for logo recognition learning \cite{zhu2007automatic, FlickrLogos, joly2009logo, psyllos2010vehicle, boia2014local}. Apart from SIFT, other approaches were also explored mostly by the information retrieval community using min-hashing \cite{romberg2013bundle}, metric learning \cite{chen2003noisy}, Vocab trees \cite{nister2006scalable}, using text \cite{sivic2003video}, bundling features for improved search \cite{wu2009bundling}. Most of these approaches needed complex image preprocessing pipelines along with several independent models. 

Recent initiative in logo recognition use deep neural networks, which offer superior performance with end to end pipeline automation, i.e. from image and logo identification to recognition. Broadly speaking, the following approach is prevalent - an image is fed to deep neural object detector and classifier gives out predictions \cite{hoi2015logo, bianco2017deep, iandola2015deeplogo, bianco2015logo, oliveira2016automatic, tuzko2017open, su2018open}, most of these approaches use an ImageNet \cite{krizhevsky2012imagenet} trained CNN which is fine-tuned on FlickrLogos-32 \cite{FlickrLogos} dataset. The lack of a big quality logo dataset makes the models less generalizable. Other datasets like WebLogo-2M \cite{su2018weblogo} are large but this dataset is noisy with no manual bounding box annotations. Instead, the data is annotated via an unsupervised process, meaning the error rate is unknown. The Logos in the wild dataset is much better \cite{tuzko2017open} but it lacks the large intra- and inter-class variations that \textit{PL2K} provides. 

\subsection {Metric learning}
Distance Metric learning (DML) has a very rich research history in information retrieval, machine learning, deep learning and recommender systems communities. DML has successfully been used in clustering \cite{hershey2016deep}, near duplicate detection \cite{zheng2016improving}, zero-shot learning \cite{oh2016deep}, image retrieval \cite{sohn2016improved}. We briefly cover it in relation to our work (deep methods). The seminal work in DML is to train a Siamese network with contrastive loss \cite{hadsell2006dimensionality, chopra2005learning}, where pairwise distance is minimized for image pairs with same class label and push distance between dissimilar pairs greater than some fixed margin. A downside of this approach is that it focuses on absolute distances, whereas for most tasks, relative distances matter more. One improvement over contrastive loss is triplet loss \cite{schultz2004learning, weinberger2009distance} which constructs a set of triplets, where each triplet has an anchor, a positive, and a negative example where anchor and positive have the same class label and negative has a different class label.

In practice, the performance of these methods heavily depends on pair mining strategies as there are exponential number of pairs (mostly negative pairs) that can be generated. Several works in recent years have explored various smart pair mining strategies. Facenet \cite{schroff2015facenet} proposed online hard negative mining strategy but this technique has a short-coming where it empirically required larger batch sizes to work. Curriculum learning \cite{bengio2009curriculum} was explored by \cite{appalaraju2017image} where a probability distribution was used to sample image pairs online in order of their hardness, with easy to cluster image pairs sampled more earlier on in training and harder to identify pairs introduced later on in training. There exists other DML sampling approaches trying to devise losses easier to minimize using small mini-batches \cite{oh2016deep, sohn2016improved, hershey2016deep, ionescu2015training, wang2014learning, wu2017sampling}. 

Relating this prior art with our contributions in logo research; none of these approaches explore logo recognition as a few-shot clustering formulation which we feel is a better fit for this problem. As a result, we did not have to perform class imbalance correction (as done by \cite{su2018weblogo}), our approach can handle large number of logo classes and do effective few-shot logo detection by projecting new logos into an embedding space. We used a combination of triplet-loss and proxies\cite{proxyloss} to optimize this embedding space and not a simple distance measure. We hypothesize that the principle of proposed method can be applied effortlessly to other image tasks like classification or object detection. We picked logo recognition due to its wide range of applications and deep research history. 

%-------------------------------------------------------------------------
\section{Approach}

As discussed earlier, one of the key challenges with logo detection is that the context in which the logo is embedded can vary almost infinitely. State of the art deep learning object detectors that are trained to localize and identify a closed set of logos will inherently use the contexts of each logo for training and prediction which makes them susceptible to context changes. For example, a logo that appears only on shoes in the training data might remain invisible or get confused by same logo if it is displayed on a coffee mug during inference.

To overcome this issue, we propose a two-step approach, where first a semantic logo detector identifies rectangular regions of an image where a logo might be located and a second model, logo recognizer identifies its class/brand. In contrast to recent works \cite{tuzko2017open} that apply state of the art object detectors such as Faster R-CNN \cite{fasterrcnn} or SSD\cite{liu2016ssd} to detect and identify a fixed set of logos, we aim at a universal logo detector that does not need further retraining if the classes change. This method also alleviates the problem of collecting and annotating a new large body of training data for every future logo that needs to be detected.

Given a large number of training images across a wide range of brands and contexts, we expect the models to learn the abstract concept of \emph{logoness} and to be able to work with any logo class at inference time. In practice, we train these models in a class-agnostic way: every generated region proposal is classified in a binary fashion: logo or background discarding any class-related information. In Section \ref{sec:experiments} we discuss results of this claim of universal logo detection on \textit{PL2K} dataset and on public logo dataset FlickrLogos-32. We also run different state-of-the-art object detector architectures (SSD, Faster R-CNN, YOLOv3) and analyze their results. 

\subsection{Few-shot Logo Recognition}

Once the semantic logo detector has identified a set of probably logo regions within a given image we need to have a mechanism that can correctly classify these regions into its corresponding logo class/brand. Ideally, this step could be solved via a state of the art CNN image classification model such as ResNets\cite{he2016deep} with multi-class classification. However this necessitates the right amount of training data for every class, class imbalance corrections and might also constrain the number of classes.

Recent advances in deep embedding learning propelled the research in few- \cite{xian2018zero, kimurafew} or one- \cite{santoro2016one} or zero-shot learning \cite{lei2015predicting} where the aim is to use only a few, single or no examples of each class during training. The typical way this is achieved is via metric learning, where a model learns the similarity among arbitrary groups of data, thus being able to cope even with a large number of (unseen) classes.

Currently, state-of-the-art methods for metric learning employ deep (convolution) neural networks, which are trained to output an embedding vector for each input image so that it minimizes a loss function defined over the distances of points. Usually, distances are learned using triplets of similar and dissimilar points $D = {(x, y, z)}$, where $x$ being the anchor, $y$ the positive, and $z$ the negative point and $d$ is the Euclidean distance function. With $y$ being more similar to $x$ than $z$ the task is to learn a distance respecting the similarity relationships encoded in D:

\begin{equation}
d(x,y) \le d(x,z) \text{ for all } (x,y,z) \in D
\end{equation}

\begin{figure}[t]
	\begin{center}
		\includegraphics[width=\linewidth]{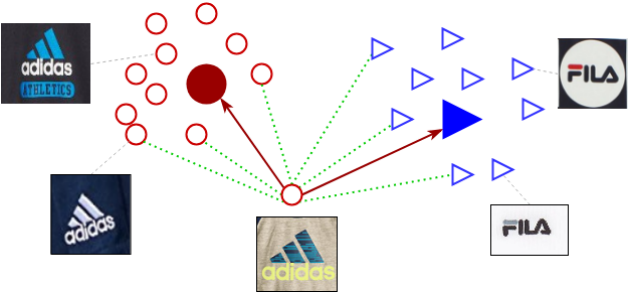}
	\end{center}
	\caption{Example of proxies used during training: for the Adidas logo (in the middle) distances are computed to the positive (dark red) proxy to its left and negative (blue) proxy to its right instead of other training images (small circles / triangles). Proxies do not belong to any single image, they are learned during training.}
	\label{fig:proxies_example}
\end{figure}

Triplet-loss addresses this with a hinge function to create a fixed margin between the anchor-positive difference, and the anchor-negative difference:
\begin{equation}
L_{triplet}(x, y, z) = [d(x,y) + M  - d(x,z)]_{+}
\end{equation}
However, it has been shown that the performance of these functions depends greatly on the way these pairs and triplets are sampled \cite{schroff2015facenet, appalaraju2017image}. In fact, computing the right set of triplets is a computationally expensive task which has to be performed for every mini-batch during training for optimal results. Movshovitz-Attias \etal introduced the notion of proxies \cite{proxyloss} in combination with \textit{NCA} loss \cite{NCALoss} that completely removes the sampling burden while providing state-of-the-art performance on CUB200 \cite{wah2011caltech}, Cars196 \cite{krause20133d} and Stanford Products\cite{oh2016deep} datasets. They \cite{proxyloss} define \textit{NCA} loss over proxies the following way:
\begin{equation}
L_{NCA}(x,y,Z) = - \log \bigg(\frac{\exp(-d(x,p(y)))}{    \sum_{z\in Z}\exp(-d(x,p(z)))     }\bigg)\text{,}
\end{equation}
where $Z$ is a set of all negative points for $x$ and $p(x)$ is the \emph{proxy} for $x$ with $p(x) = \arg \min d(x,p)$
for all $p \in P$ that we need to learn. See Figure \ref{fig:proxies_example} for an illustrative example of proxies. Similar to the original work we train our model and all proxies with the same norm: $N_{p} = N_{x}$. Since \textit{NCA} loss even with proxies over-fit very early for our logo identification problem, we used the original triplet loss with proxies:
\begin{equation}
L_{triplet}(x, y, Z) = [d(x,p(y)) + M  - d(x,p(Z))]_{+}\text{,}
\end{equation}
We choose one proxy per logo class with static proxy assignment. This yields fast convergence and state-of-the-art results on FlickrLogos-32 without using any triplet-sampling strategy. In the experiments section we show how this simple change in the loss function leads to superior performance without sampling. 

At inference time, using this trained embedding function, one can apply (approximate) K-nearest neighbor search among embedding vectors to find the corresponding prediction for each query image.

\section{Product Logo Dataset (PL2K)}

\begin{table*}[h]
	\centering
	\begin{tabular}{c||c|c|c|c|c|c}
		Dataset & Logos & Images & Supervision & Noisy & Construction & Scalability \\
		\hline
		\hline
		BelgaLogos \cite{BelgaLogos} & 37 & 10,000 & Object-Level & \xmark & Manually & Weak \\
		\hline
		FlickrLogos-32 \cite{FlickrLogos} & 32 & 8,240 & Object-Level & \xmark & Manually & Weak \\
		\hline
		FlickrLogos-47 \cite{FlickrLogos} & 47 & 8,240 & Object-Level & \xmark & Manually & Weak \\
		\hline
		TopLogo-10 \cite{TopLogo10} & 10 & 700 & Object-Level & \xmark & Manually & Weak \\
		\hline
		Logo-NET \cite{LogoNet} & 160 & 73,414 & Object-Level & \xmark & Manually & Weak \\
		\hline
		WebLogo-2M \cite{su2018weblogo} & 194 & 1,867,177 & Image-Level & \cmark & Automatically & Strong \\
		\hline
		Logos in the wild \cite{tuzko2017open} & 871 & 11,054 & Object-Level & \cmark & Manually & Medium \\
		\hline
		\textbf{PL2K (Ours)} & 2000 & 295,814 & Object-Level & \xmark & Semi-automatically & Strong \\
		\hline
		
	\end{tabular}
	\caption{Statistics and characteristics of existing logo detection datasets.}
	\label{tab:datasettable}
\end{table*}

Ideally, our training set would be a large body of annotated images featuring logos across a high variety of contexts and domains to be able to train a deep-learning based class-agnostic object detector model as these models have millions of trainable parameters. Unfortunately, no current public dataset satisfies these requirements. One reason could be that image collection and annotation for such a dataset would require expensive manual work to keep quality high and to prevent potential copyright issues that might arise when images are farmed with automated scripts from the Internet. Also care has to be taken to not annotate counterfeit logos as we would not want the models to learn the wrong logo representation.

Therefore, we decided to build a new dataset by sampling images from the Amazon Product Catalog for the following reasons:
\begin{itemize}
	\item A large body of publicly accessible images are available, thus easy to automate data collection.
	\item Images are labeled with the corresponding brand that helps with annotation.
	\item We are interested in the abstraction capacity of object detection models i.e. how they perform on non-product images.
	%\item describe the dataset more. Ideas: \textit{WebLogo-2M}: Scalable Logo Detection by Deep Learning from the Web \cite{su2018weblogo}
\end{itemize}

Product images on Amazon typically feature a single or multiple products in front of a white background. Our working hypothesis is that a large amount of product images will offer a high enough variance in logo contexts from which the object detection model can learn and generalize the concept of a Logo. Furthermore, we chose 2000 brands based on popularity that satisfy the following conditions: 
\begin{enumerate*}[label=(\roman*)]
	\item have a significant number of images
	\item well-established logo
	\item logo frequently used on the product.
\end{enumerate*}

We sampled a total of 1 million product images randomly from these brands and used Amazon Mechanical Turk to annotate them. Every image was sent to 9 different \textit{MTruk} workers for annotation. Each worker had to complete the following task:
\begin{enumerate}
	\item Identify if the image contains no, one or multi logos; label image as such \emph{NO LOGO, ONE LOGO, MULTIPLE LOGO}.
	\item For the bounding box: in a separate task, of the images with at least one logo, the workers were instructed to locate the leftmost, topmost logo on the image (if multiple present) and draw a rectangle around it. If there were still multiple options, the workers were instructed to choose the biggest one.
\end{enumerate}

% Post-processing
\textbf{Post-processing:} Due to the different interpretations of the term logo, we received very different annotations for the same image from multiple workers. To consolidate the results, we filtered out every image marked as \emph{NO LOGO} by more than 3 workers (out of 9). This finally gave us 295,814 images, each with at least 6 bounding-box annotations.

In order to reduce noise and accurately merge the annotation box we used the DBScan clustering algorithm \cite{Ester96adensity-based} as it requires no \emph{a priori} knowledge on the number of clusters. The algorithm was performed on a precomputed pairwise distance matrix of the annotation rectangles where the distance was defined as the complement of intersection over union (IoU), see equation \ref{eq:iou_complement}. We empirically derived epsilon to be 0.6 and the minimum core samples to be 1 as these yielded the best results.
\begin{equation}
D = 1- \frac{R_{1} \cap R_{2} }{R_{1} \cup R_{2}}\text{.}
\label{eq:iou_complement}
\end{equation}
As we started using \textit{PL2K}, we found that several annotators marked the whole image as a logo even though that was clearly incorrect. Therefore, we removed all merged bounding boxes that have an $IoU > 0.65$ with the image rectangle. We removed nearly 36k bounding boxes and about 2K images from the dataset. Finally, we split the dataset into training plus validation (80\%) and testing (20\%) making sure that sets do not share images and brands taken from the same set. Ultimately, we are interested in the generalization capability of the model on unseen brands.

\begin{table}
	\begin{center}
		\begin{tabular}{ c|c|c }
			Image set & No. of images & No. of brands \\ 
			\hline
			Training & 185247 & \multirow{2}{*}{206} \\ 
			Validation & 46312 & \\ 
			\hline
			Testing & 57970 & 1528 \\ 
			\hline
			Negatives & 10000 & 2000 \\ 
		\end{tabular}
	\end{center}
	\caption{PL2K data split for train and test.}
	\label{tab:PL2LKDataset}
\end{table}

We also added a negative set that consists of randomly sampled images that were marked as containing no logos by all annotators. We uses this set to measure the false positive rates of the models. Table \ref{tab:PL2LKDataset} provides a quick overview of the \textit{PL2K} dataset while Table \ref{tab:datasettable} compares \textit{PL2K} with existing logo detection datasets. Note, that we split the data in such a way that there are almost 8 times more logos in test data than in train data. We wanted to showcase the universal object detector's generalization capabilities.\newline
% Identification set
\textbf{Data for Few-shot logo detector:} The second part of our data collection effort was for few-shot logo detector. This was slightly different from universal logo detector. This model operates on logo regions i.e. logo appearances cropped out from images. We picked product images for the top 242 logos and ran them through our universal logo detector. Based on the regions proposed by universal logo detector, we manually filtered out false positive regions and identified 242 valid brand logos. From this we had at least 700 cropped regions for each logo. This dataset was then split into a train and test set (80/20\%) with 193 and 49 logo classes respectively.

\section{Experiments}
\label{sec:experiments}

We split our experiments into two parts: first we investigate the performance of the Universal logo detector operating on \textit{PL2K} and FlickrLogos-32. As discussed in table \ref{tab:datasettable}, \textit{PL2K} is relatively clean Amazon catalog images and FlickrLogos-32 are real-world logos in the wild dataset. Second, we discuss our experiments with the few-shot logo classifier which works with cropped image regions. Finally in end-to-end section, we see how these techniques perform on FlickrLogos-32 without fine-tuning on this dataset.

\subsection{Universal Logo Detection}

The following three state-of-the-art object detector architectures with two output classes were tested for our universal logo detector: Faster R-CNN \cite{fasterrcnn}, SSD \cite{liu2016ssd}, and YOLOv3 \cite{yolov3}. Faster R-CNN is a two step detector that has a higher performance on standard datasets (e.g. MS COCO\cite{mscoco}), but is a lot slower than the other two single-shot detectors. There have been several proposals on improving the performance of this model \cite{focalloss, FPN} but these only offer a minor increase in mAP at the cost of speed.

\begin{figure}[t]
	\begin{center}
		\includegraphics[width=\linewidth]{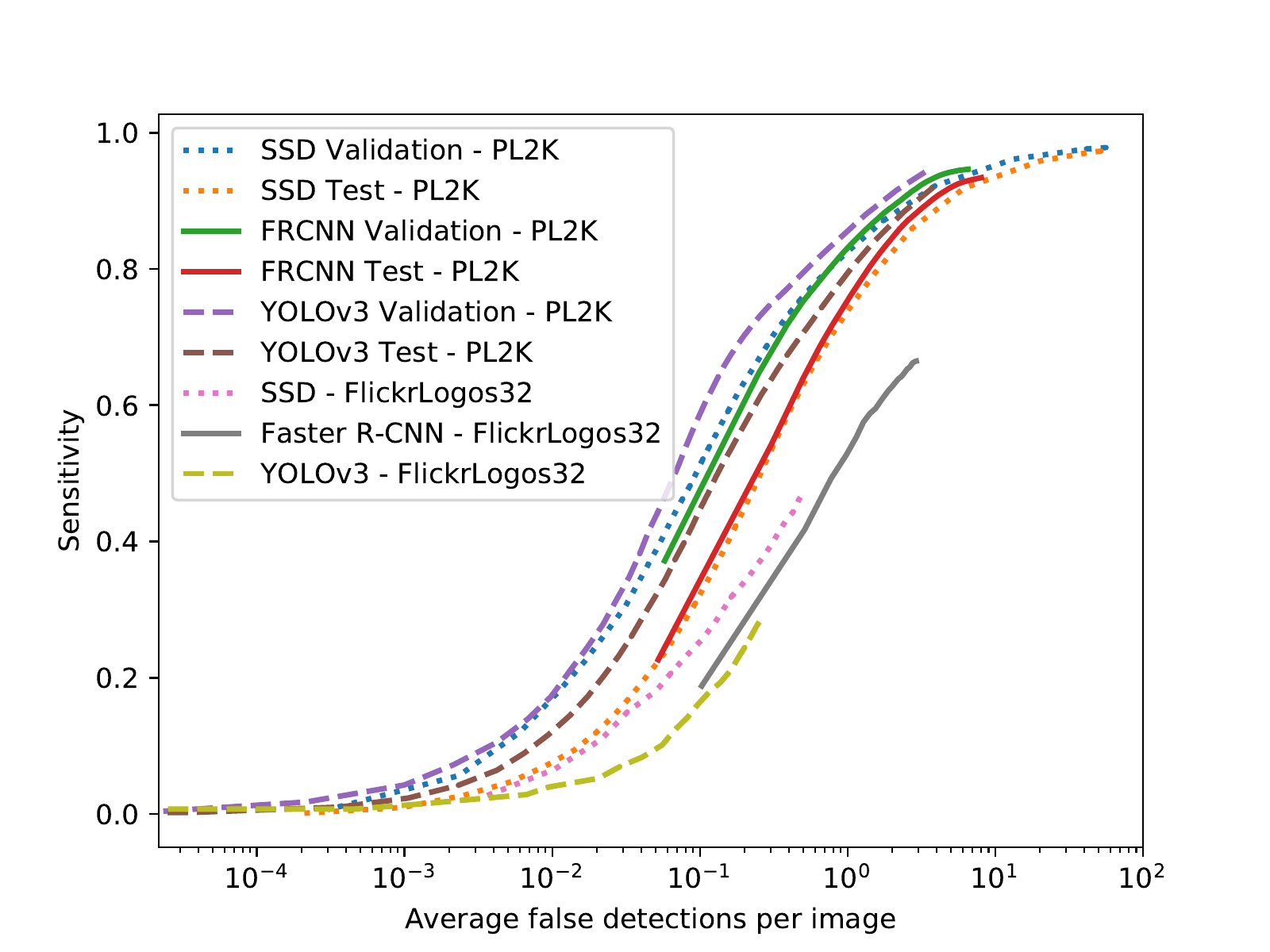}
	\end{center}
	\caption{FROC curves of the three detector models on FlickrLogos-32. Faster R-CNN achieves higher recall but with lot more false positives. Best viewed in color.}
	\label{fig:froc_curves1}
\end{figure}

We used a ResNet50 \cite{he2016deep} base CNN for all networks except for \textit{YOLOv3} which uses the Darknet53 architecture. All of the networks were pre-trained on ImageNet \cite{deng2009imagenet} and then trained end-to-end on the \textit{PL2K} dataset with a fixed input size of 512x512 for 20 epochs. YOLOv3 was trained with randomly resized inputs in the range of [320, 640] with steps of 32, to achieve better accuracy. We computed the recall at $IoU > 0.5$, average precision (AP), and the number of regions generated on the negative/no logo set.\newline
\textbf{Same domain (PL2K).} We find that all models achieve very high recall and AP values with SSD having the highest recall and YOLOv3 the highest AP on the \textit{PL2K} validation dataset. This trend repeats on the test set with a 0.1 drop in AP and 1-2\% points in recall meaning the models still perform well on a wide range of unseen logos and products. Interestingly, the two-step detection by Faster R-CNN does not bring any advantage in performance. On the contrary, it increases the false positive rate by a large margin. The FROC curves on Figure \ref{fig:froc_curves1} depict this behavior accurately.\newline
\textbf{Different domain (FlickrLogos-32).} In order to see how these models work on a completely different domain without fine-tuning we ran the same evaluation on the FlickrLogos-32 dataset. This dataset is the most popular evaluation dataset for logos, consisting of 8,240 images covering 32 logos/brands. The performance trend seemed to reverse: Faster R-CNN with an accuracy of close to 80\% and AP of 0.42 outperforms SSD and YOLOv3 by a large margin. This suggests that Faster R-CNN learned less domain-specific features which combined with the almost 8x more predictions outperforms all other open-set detectors reported by \cite{tuzko2017open}. See Table \ref{tab:detectionresults} for the full set of results and Figure \ref{fig:froc_curves1} for the corresponding FROC curves.

Based on these experiments, we chose Faster R-CNN as it has superior generalization capabilities. This model is what helped achieve the state-of-the-art results on FlickrLogos-32 (see table \ref{tab:universallogodetectperf}). More analysis revealed that SSD and YOLOv3 had issues detecting smaller bounding box regions, images with occlusions or logo which blend into their environment. 

\begin{table*}[htp]
	\centering
	\begin{tabular}{c||c|c|c|c|c|c|c|c}
		\multirow{2}{*}{Model} & \multicolumn{2}{c|}{Validation set} & \multicolumn{2}{c|}{Test set} & Negative set & \multicolumn{3}{c}{FlickrLogos-32} \\
		& Recall & AP & Recall & AP & No. detections & Recall & AP & Negative detections \\
		\hline
		\hline
		Faster R-CNN & 94.76\% & 0.72 & 93.52\% & 0.63 & 147152 & \textbf{79.87\%} & \textbf{0.42} & 8379  \\
		\hline
		SSD	& \textbf{98.05\%} & 0.73 &	\textbf{97.73\%} & 0.62 & \textbf{19295} & 60.04\% & 0.38 & 1136 \\
		\hline
		YOLOv3 & 94.29\% & \textbf{0.77} & 92.10\% & \textbf{0.70} & 19504 & 44.69\% & 0.22 & \textbf{985}	 \\
	\end{tabular}
	\caption{Universal Logo Detector performance on the \textit{PL2K} and FlickrLogos-32 datasets.}
	\label{tab:detectionresults}
\end{table*}

\subsection{Few-shot Logo Identification}

For the few-shot logo embedding model we used the SE-Resnet50 \cite{hu2017squeeze} architecture with the same modifications as described in\cite{arcface}. Input images were resized to 160x160 pixels, the embedding dimension was 128 and the batch size 32. We used the Adam optimizer with momentum 0.9, weight decay 0.0005, and learning rate $10^{-4}$ which we reduced by a factor of 0.8 every 20 epochs. The network's parameters were initialized using Xavier initialization  \cite{glorot2010understanding} with magnitude 2, no transfer learning was used. 

\begin{figure}[t]
	\begin{center}
		\includegraphics[width=0.9\linewidth]{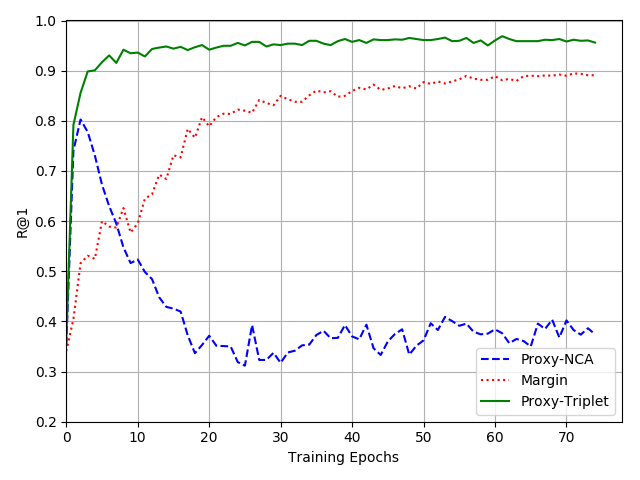}
	\end{center}
	\caption{Performance of various distance learning loss functions for Few-shot model on the \textit{PL2K} test set.}
	\label{fig:loss_performance}
\end{figure}

We trained few-shot model by passing \textit{PL2K} annotated images to the few-shot logo detector (section on \textit{PL2K} dataset). The 242 logo classes were used for training and testing without any sampling strategy. Various loss functions and spatial transformer layers (see table \ref{tab:top1RecallLossMetrics}) were tried on top of this base architecture. 

For comparison, we trained the same model using distance-weighted sampling with margin-based loss\cite{wu2017sampling} as well as proxy-NCA loss since both methods report higher performance than triplet loss with various sampling strategies. As a solid baseline we also added cross-entropy loss, though here we are using the last layer of the feature extractor, thus the embedding dimension is much larger than 128.

\begin{figure}[t]
	\begin{center}
		\includegraphics[width=\linewidth]{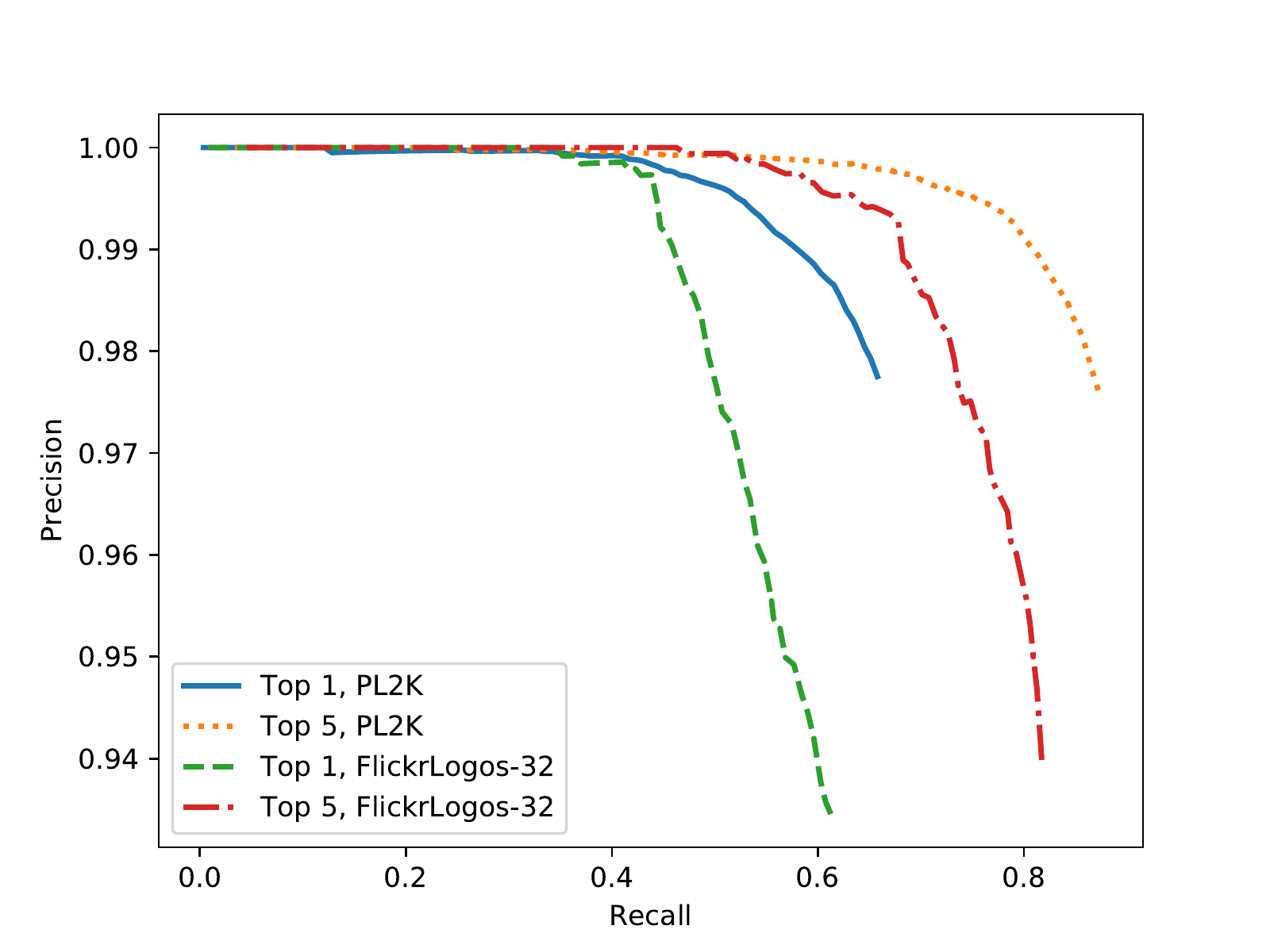}
	\end{center}
	\caption{Precision-Recall metric for few shot logo detector on \textit{PL2K} and FlickrLogos-32 data. Note the state-of-the-art performance of the few-shot logo detector on Flickr-32 data. The model has not seen FlickrLogos-32 data in training. Best viewed in color.}
	\label{fig:PrecisionRecall}
\end{figure}

As seen on Figure \ref{fig:loss_performance} the proxy-triplet loss converges very fast to a superior score in contrast with the other approaches. For the first few epochs proxy-NCA has almost identical performance then it starts to diverge and quickly decline (over-fit). This suggests that proxy-augmented loss functions are strongly dataset and/or hyper-parameter sensitive but it needs further investigation.

\begin{figure}[t]
	\begin{center}
		\includegraphics[width=\linewidth]{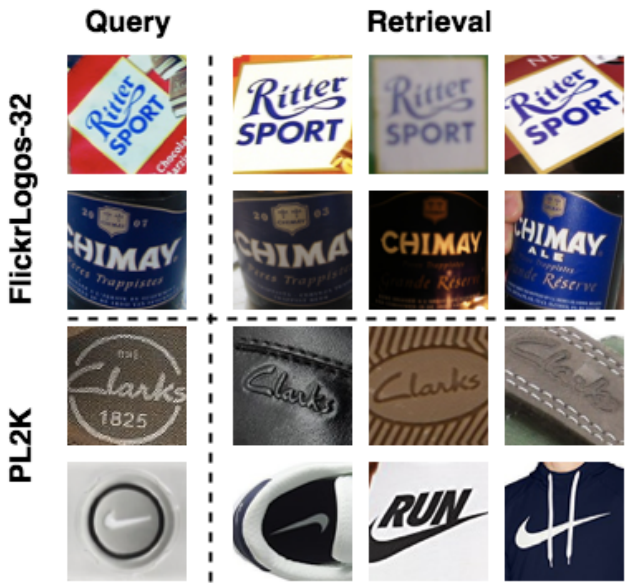}
	\end{center}
	\caption{Retrieval results of the few-shot model on the FlickrLogos-32 and \textit{PL2K} datasets. Left column contains query regions.}
	\label{fig:few_shot_retrieval}
\end{figure}

Most logos do include words and other characteristics which could provide some hint about the right orientation of a logo in noisy contexts, we also experimented with adding a Spatial Transformer network (STN) layer \cite{STN} to the proxy-triplet model, which improved the accuracy by a small margin. See the table \ref{tab:top1RecallLossMetrics} below for the best top 1 recall scores. Fig. \ref{fig:PrecisionRecall} shows the performance of the final few-shot model on \textit{PL2K} and FlickrLogos-32 dataset.

\begin{table}
	\begin{center}
		\begin{tabular}{ c|c } 
			\textbf{Few-shot Logo model} & \textbf{Top 1 Recall} \\ 
			\hline
			CrossEntropy Loss & 89.10\% \\
			\hline
			Margin Loss & 91.09\% \\
			\hline
			Proxy-NCA Loss & 80.26\% \\
			\hline
			Proxy-Triplet Loss & 96.80\% \\
			\hline
			Proxy-Triplet Loss with STN & \textbf{97.16\%} \\
		\end{tabular}
		\caption{Top1 recall of few-shot Resnet50 model and various losses on \textit{PL2K} dataset.}
		\label{tab:top1RecallLossMetrics}
	\end{center}
\end{table}

\subsection{Qualitative Analysis}
\begin{figure}[t]
	\begin{center}
		\includegraphics[width=\linewidth]{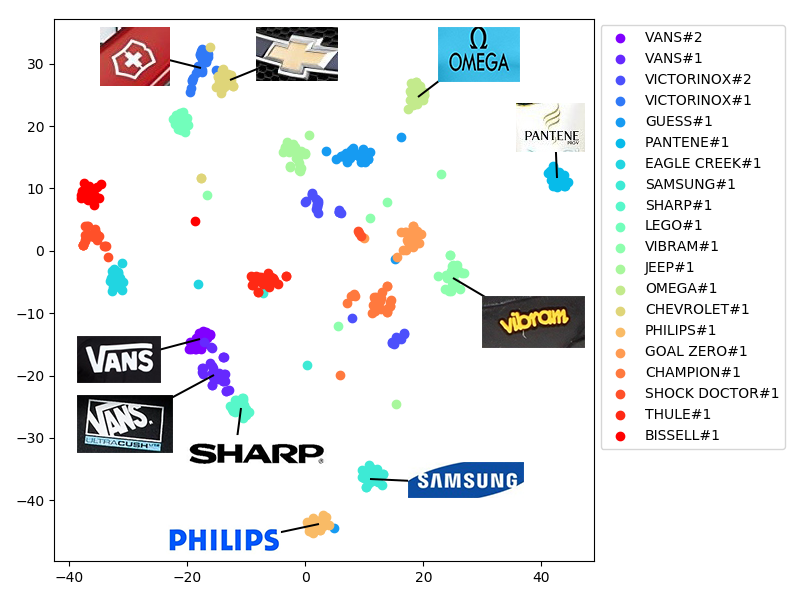}
	\end{center}
	\caption{t-SNE \cite {maaten2008visualizing} plot of a random subset of test classes. The model successfully maps logos of the same class close to each other, even with high inter- and intra-class variations (e.g. VANS\#1-\#2, or Samsung and Philips). Best viewed in color.}
	\label{fig:tsne_plot}
\end{figure}

To get a better understanding on quality of the trained solution we ran the t-SNE algorithm \cite {maaten2008visualizing} on a randomly selected subset of test classes for 1000 iterations with perplexity 40 (see Figure \ref{fig:tsne_plot}). We find that our model successfully separates very similar looking logo classes even if they use similar font or color. There are a few single points scattered around the space (in the middle) these are impressions from logos that are radically different in the use of color, shape, font than the rest of the logos.

\begin{figure}[t]
	\centering
	\setlength\cellspacetoplimit{4pt}
	\setlength\cellspacebottomlimit{4pt}
	\begin{tabular}{ Sc Sc}
		Query Image & Nearest neighbor  \\
		\hline
		\includegraphics[height=0.2\linewidth]{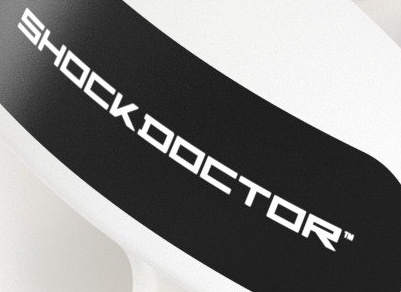} & \includegraphics[height=0.2\linewidth]{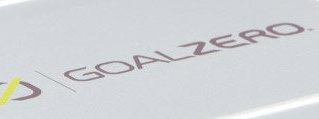} \\
		
		\includegraphics[height=0.2\linewidth]{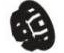} & \includegraphics[height=0.2\linewidth]{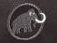} \\
		
		\includegraphics[height=0.2\linewidth]{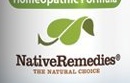} & \includegraphics[height=0.2\linewidth]{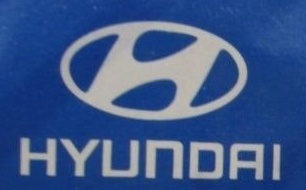} \\
	\end{tabular}
	\caption{Erroneous classifications: few-shot recognition becomes hard when image resolution is low and/or logos are made of very simple shapes}
	\label{tab:misdetects}
	\label{fig:misclassified}
\end{figure}

Figure \ref{fig:few_shot_retrieval} illustrates some of the successful retrieval results on both datasets. In Figure \ref{fig:misclassified}, we show a few examples that are wrongly classified (right column) by our model based on query image (left column). Diving deeper into the mis-classified examples suggested low resolution images, logo consisting of very simple shapes being a few reasons for mis-classification. 

\subsection{End-to-end evaluation}
We evaluate the performance of our universal detector combined with the few-shot identification model on FlickrLogos-32 a popular logo datasets and compare it to the state-of-the-art. None of the models were trained or fine-tuned on FlickrLogos-32 dataset. All models were trained on \textit{PL2K} dataset. As seen in Table \ref{tab:universallogodetectperf} and table \ref{tab:top1RecallLossMetrics}, the final model used was Faster R-CNN as universal Logo detector; few-shot model was a SE-Resnet50 with proxy-triplet loss with Spatial transformer layers. Top 5 accuracy worked best as it generated more proposals. 

For few-shot we extracted the first five ground truth regions per brand and used it as anchors. These anchor regions were excluded from evaluation to avoid bias. We also ran two evaluations per each detector type (single and five shot). Faster R-CNN worked best with mAP of 0.56558. mAP is decided by region proposals with class detection threshold of 0.5.

\begin{table}
	\begin{center}
		\begin{tabular}{ c|c|c|c } 
			End to End Model & mAP@1 & mAP@5 & \begin{tabular}[c]{@{}c@{}}No.\\proposals\end{tabular} \\  %No. proposals
			\hline
			SSD + FS* & 35.833\% & 44.79\% & 2655 \\
			\hline
			Faster R-CNN + FS* & \textbf{44.42\%} & \textbf{56.55\%} & 8786 \\
			\hline
			YOLOv3 + FS* & 23.31\% & 30.12\% & 1525 \\
			\hline
			\hline
			T\"uzk\"o et al. \cite{tuzko2017open} & \multicolumn{2}{c|}{46.4\%} & N/A \\
		\end{tabular}
	\end{center}
	\caption{Top 1 and Top5 Accuracy of the end to end evaluation on FlickrLogos-32 dataset. This uses both Universal Logo detector and Few-shot Logo recognizer. *FS=Few-shot model SE-Resnet50 with Proxy-Triplet Loss with STN as shown in table \ref{tab:top1RecallLossMetrics} (Row 1-3). Row 4 shows the results of the only reported open-set detector. Note, that this system was trained using FlickrLogos-32.}
	\label{tab:universallogodetectperf}
\end{table}

\section{Conclusion and Future work}
In this work we shared our approach to few shot logo recognition using deep learning two stage models. We trained our models on \textit{PL2K} and evaluated on FlickrLogos-32 to achieve new state-of-the-art performance of 56.55\% mAP@5. This empirically indicates that our approach does good domain adaptation without fine-tuning. 

We also conducted extensive experiments on various CNN architectures and compared with existing logo works to show that triplet-loss with proxies is an effective way to find similar images. We also presented product logo dataset \textit{PL2K}, a first of its kind large scale logo dataset.

Future extensions of this works could look at application of this work in broader contexts of image similarity search, generic object detection or going deeper into understanding why triplet-loss with proxies does so well and its limitations.

%------------------------------------------------------------------------
\section{Acknowledgments}
We would like to thank Arun Reddy and Mingwei Shen for giving us the opportunity to work on this problem. We would also like to thank Niel Lawrence, Rajeev Rastogi, Avi Saxena, Ralf Herbrich, Fedor Zhdanov, Dominique L'Eplattenier, Marc Ascolese and Ives Macedo for providing constructive feedback on our work.  Finally, Ankit Raizada for his work on putting together the \textit{PL2K} dataset and John Weresh for naming it.

{\small
	\bibliographystyle{ieee}
	\bibliography{egbib}
}

\end{document}